\documentclass{article}
\usepackage{ijcai19}

\usepackage{times}
\usepackage{soul}
\usepackage{url}
\usepackage[colorlinks=true,linkcolor=black,citecolor=black,bookmarks=true]{hyperref}
\usepackage[utf8]{inputenc}
\usepackage[small]{caption}
\usepackage{graphicx}
\usepackage{amsmath}
\usepackage{booktabs}
\usepackage{diagbox}
\usepackage{algorithm}
\usepackage{algorithmic}
\usepackage{amsfonts}
\usepackage{color,xcolor}
\usepackage{makecell}
\usepackage{rotating}
\usepackage{multirow}
\usepackage{pifont}
\usepackage{enumitem}

\usepackage{multirow} 
\usepackage{overpic}
\usepackage{amssymb}

\usepackage{wrapfig}
\usepackage{picinpar}
\usepackage{cutwin}

\usepackage{latexsym}
\usepackage{etoolbox}
\apptocmd{\sloppy}{\hbadness 10000\relax}{}{}

\usepackage{fancyhdr}

\title{LSANet: Feature Learning on Point Sets by Local Spatial Aware Layer}

\author{
Lin-Zhuo Chen\textsuperscript{*} 
\and
Xuan-Yi Li\textsuperscript{*}  \and
Deng-Ping Fan \and
Kai Wang \and
Shao-Ping Lu \And
Ming-Ming Cheng
\affiliations
College of Computer Science, Nankai University\\
}
\graphicspath{{./Imgs/}}
\DeclareGraphicsExtensions{.pdf,.png,.jpg}

\newcommand{\figref}[1]{Fig.~\ref{#1}}
\newcommand{\tabref}[1]{Tab.~\ref{#1}}
\newcommand{\secref}[1]{Sec.~\ref{#1}}

\newcommand{\equref}[1]{Equ. (\ref{#1})}

\def\eg{\emph{e.g.~}}

\def\sArt{{state-of-the-art~}}

\begin{document}

\maketitle

\begin{abstract}
Directly learning features from the point cloud 
has become an active research direction in 3D 
understanding. 
Existing learning-based methods usually construct 
local regions from the point cloud and extract 
the corresponding features. 
However, most of these processes do not adequately 
take the spatial distribution of the point 
cloud into account, limiting the ability to 
perceive fine-grained patterns. 
We design a novel \emph{Local Spatial Aware (LSA)} layer, 
which can learn to generate Spatial Distribution Weights (SDWs)
hierarchically based on the spatial relationship in local region 
for spatial independent operations, to 
establish the relationship between these 
operations and spatial distribution, thus capturing 
the local geometric structure sensitively.
We further propose the \emph{LSANet}, which is based 
on LSA layer, 
aggregating the spatial information with 
associated features in each layer of the 
network better in network design.
The experiments show that 
our \emph{LSANet} can achieve on par or better 
performance than the state-of-the-art methods 
when evaluating on the challenging benchmark datasets. For example, 
our \emph{LSANet} can achieve 93.2\% accuracy on ModelNet40
dataset using only 1024 points, 
significantly higher than other methods under the same conditions.  The source code is available
at \url{https://github.com/LinZhuoChen/LSANet}.
\end{abstract}

\vspace{-4pt}
\section{Introduction}
\label{sec:section1}
\vspace{-4pt}
\let\thefootnote\relax\footnotetext{* Equal contribution}
With the rapid growth of various 3D sensors, how to effectively 
understand the 3D point cloud data captured from those 3D sensors 
is becoming a fundamental requirement. 
In the 2D image processing domain, deep convolutional neural 
network (CNN) based methods have achieved great success in 
almost all computer vision tasks.
Unfortunately, 
it is still tricky to directly migrate these CNN based techniques 
to 3D point sets oriented research.
Point sets have their unique property of invariance to permutations 
and cannot be accurately represented by regular lattices, making 
those successful methods in 2D image domain unsuitable to be applied 
in 3D cases.
The most common direction is transforming 3D data 
to voxel grids~\cite{voxnet,octnet,ocnn,vote3deep,graham20173d} 
or multiple views of 2D images~\cite{multi-view} to take advantage of 
existing operations used in 2D images.
However, it would lead to some negative issues such as quantization 
artifacts and inefficient computation~\cite{pointnet,sonet}.

\begin{figure}[t!]
\centering
    \begin{overpic}[width=1.\columnwidth]{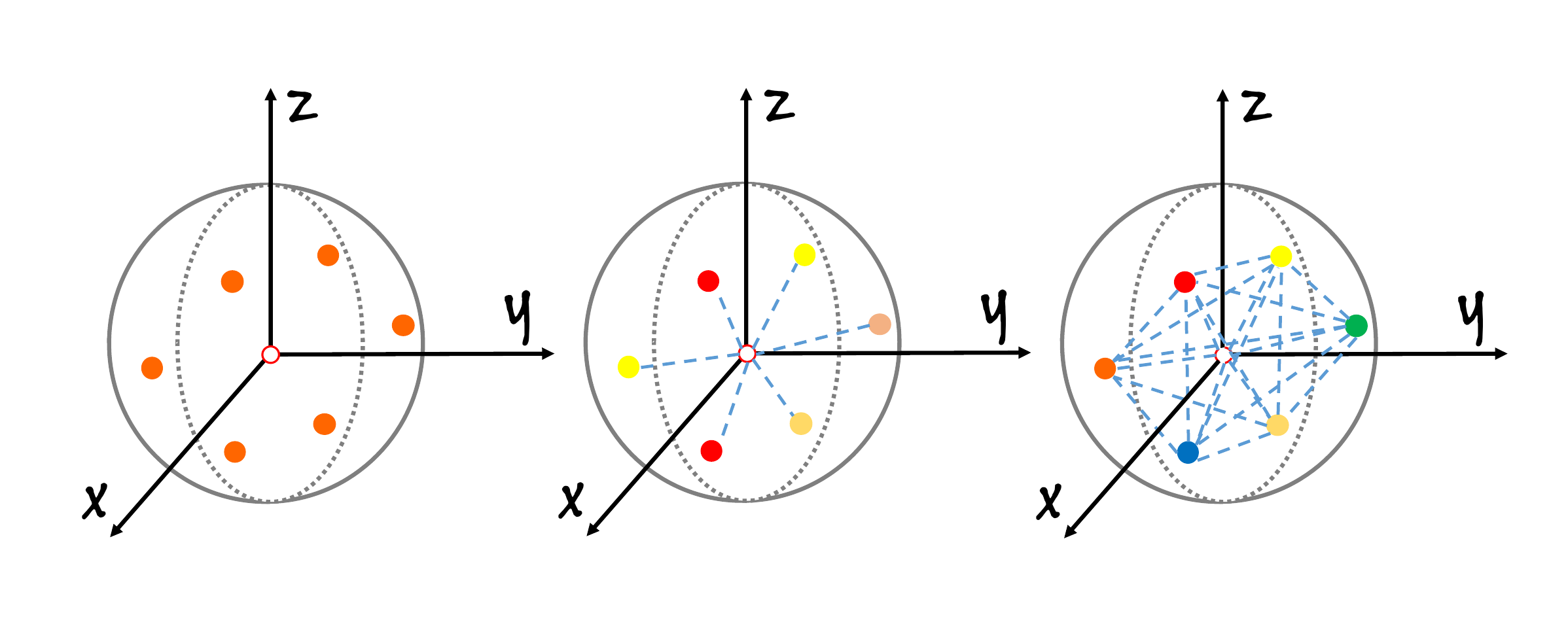}
    \put(15,2){(a)}
    \put(46,2){(b)}
    \put(76,2){(c)}
    \end{overpic}
    \vspace{-20pt}
    \caption{
    we illustrate the feature learning process for the target point (hollow circle) 
    using its neighbor points (colored circles).
    (a), (b), and (c) are three identical local regions. 
    Different colors of points represent different weights.
    (a) shows the feature extraction process of PointNet++~\protect\cite{pointnet++}:
    the weight of each point is fixed and independent 
    with the spatial information,
    making it limited to extract geometric patterns.
    (b) shows the feature extraction process of SpiderCNN~\protect\cite{spidercnn}: 
    the weight of each point is related to the vector to the center point, 
    while it does not fully consider the spatial distribution of the whole region, 
    leading to sensitivity to spatial transformation.
    (c) is an example operation in our new \emph{LSA} Layer: we can integrate our spatially-oriented SDWs into a shared MLP. 
     In our model, the weight of each point is related to all points in the local region, 
     so it can capture the local geometric structure sufficiently and obtain much robustness to geometric transform.
    }
    \label{fig:compare}
    
\end{figure}

\begin{figure*}[htbp]
    \centering
    \includegraphics[width=17cm]{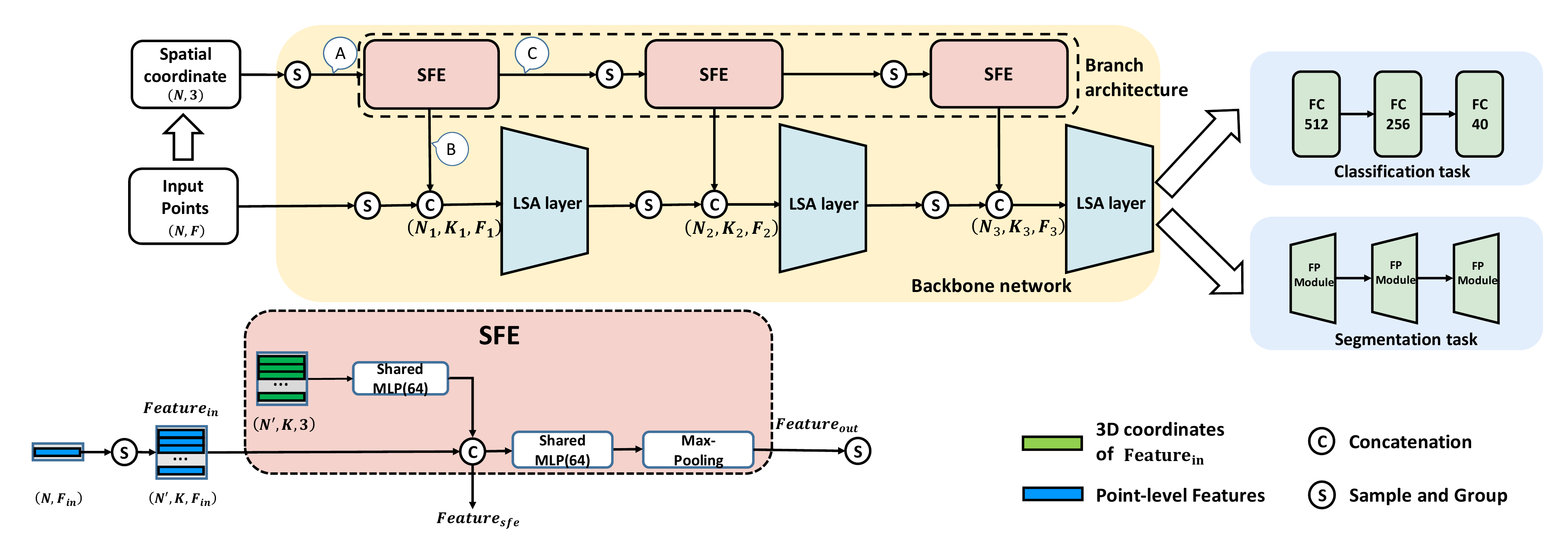}
    \vspace{-20pt}
    \caption{The architecture of LSANet for classification and segmentation: the backbone of our network is composed of \emph{SFE} and \emph{LSA} layers. The \emph{LSA} layer, which consists of SDWs generator and feature learning process, generates SDWs according to the spatial structure of each local region and integrates them with feature learning process. The details of \emph{LSA} layer are shown in \figref{fig:attention_sa_module}. For \emph{SFE}, 
    we sample and group the spatial coordinates as input which is shown in A, lift dimensions of the coordinates as output shown in B. The spatial feature as shown in C also flows into the next \emph{SFE} for a hierarchical feature representation. 
    $N_i$ represents the number of points, $K_i$ denotes the number of points in the local region, $F_i$ indicates the dimension of each point feature, and $i$ is the index of the \emph{LSA} layer.
    }
    \label{fig:pipline}
    \vspace{-10pt}
\end{figure*}

Recently, some seminal researches attempted to process point 
cloud data directly by developing specific deep learning methods, 
\eg PointNet~\cite{pointnet} and PointNet++ ~\cite{pointnet++}.
As a pioneering work, PointNet introduces a simple yet efficient 
network-based architecture, while its feature extraction is 
point-wise and thus cannot exploit the local region information.
PointNet++~\cite{pointnet++} gets the local regions 
by using farthest point sampling (FPS) and ball query algorithms, 
then extracts the features of each local region, achieving excellent results 
on different 3D datasets.
However, 
the feature extraction operations in PointNet++, which are
completed by
shared Multi-Layer Perceptron (MLP) and max-pooling, are independent with spatial structure
in the local region 
thus
can not capture the geometric pattern explicitly as 
shown in \figref{fig:compare} (a).
To overcome this difficulty,
SpiderCNN 
uses
a complicated family of parametrized non-linear
functions, where the parameters of convolution are determined 
according to the spatial coordinates in the local region. 
However, these operations only consider the spatial information 
of the single point, instead of the entire spatial distribution 
of the local region as shown in \figref{fig:compare} (b), 
thus dealing with geometric transform poorly.
%
Moreover, in PointNet++ and its improved versions, 
the raw spatial coordinates of points, 
which are relative to their center point in the local region,  
are concatenated with features of points in each layer of 
networks to alleviate the limitations of per-point operation.
However, the local coordinates have a different dimension 
and representation from the associated features. PointCNN~\cite{pointcnn}
alleviates this problem by lifting them into higher 
dimension and more abstract representation. 
In this way, with the deepening of the network, 
semantic information is gradually enriched in associated 
features, but fixed in coordinates.
%

In this paper, we propose a new network layer, named 
\emph{Local Spatial Aware (LSA)} Layer, 
to model geometric structure in local region accurately and robustly.
Each feature extracting operation in LSA layer is related to 
Spatial Distribution Weights (SDWs), which are learned based on the 
spatial distribution in local region, to establish a strong link 
with inherent geometric shape.
%
As a result, these processes can consider the local spatial distribution
as shown in \figref{fig:compare} (c), thus perceiving fine-grained 
shape patterns. 
We have also solve the problem of fixed semantic information of coordinates 
with the deepening of the network by using hierarchical Spatial Feature Extractor (SFE).
Our new network architecture, named LSANet, 
which is composed of LSA layer,
is shown in ~\figref{fig:pipline}. 
In summary, our contributions are as follows:
\begin{itemize}
\item
A novel \emph{Local Spatial Aware (LSA)} Layer 
is proposed, it establishes the relationship between
operation and spatial distribution by SDWs,
which can capture geometric structures more accurately and robustly.
%
\item
Our LSANet, taking \emph{LSA} layer as its basic unit, considers the better 
integration of space coordinates and middle layer features in design, 
and achieves the state-of-the-art results on benchmark datasets.
\end{itemize}

Extensive experiments show that the performance of our LSANet is better than \sArt methods. 
We further explain the details of the proposed \emph{LSA} layer 
and the network structure explicitly in \secref{sec:section3}.
Our results on multiple challenging datasets and ablation 
study are shown in \secref{sec:section4}.




\begin{figure*}[t]
    \centering
    \begin{overpic}[width=18cm]{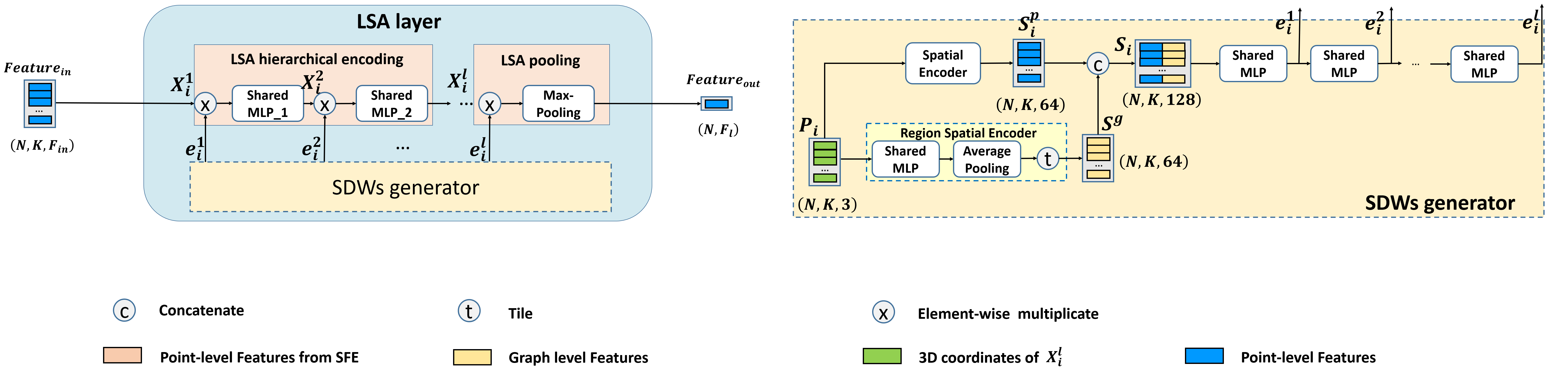}
    \put(23,6){(a)}
    \put(73,6){(b)}
    \end{overpic}
    \vspace{-10pt}
    \caption{
    {\bfseries The process of \emph{LSA} layer:}
    (a) shows the architecture of \emph{LSA} layer. (b) shows the details of SDWs generator. The \emph{LSA} layer is composed of 
    SDWs generator and the spatial independent feature learning process.
    $N$ represents the number of points, $K$ denotes the number of points in the local region, $F$ indicates the dimension of each point feature.
    }
    \label{fig:attention_sa_module}
    
    \vspace{-15pt}
\end{figure*}

\section{Related Work}
\label{sec:section2}

{\bfseries Volumetric and Multi-view approach:}
Volumetric approach converts the point sets to a regular 3D grid 
where the 3D convolution can be applied~\cite{voxnet,volumetric1}.
However, the 3D convolution usually introduces 
high computation cost, and the volumetric representations 
are often inefficient due to the sparse property of 
the point sets.
Some existing works
~\cite{octnet,ocnn,vote3deep,graham20173d,kdtree} 
aim at improving computational performance.
For instance, some representations for deep learning 
with sparse 3D data are proposed such as Octree
~\cite{ocnn}, 
Kd-Tree~\cite{kdtree}.
In~\cite{vote3deep}, the authors use a feature-centric voting scheme 
to implement a fast 3D convolution.
While in~\cite{graham20173d}, a new sparse convolutional operation 
is introduced to perform efficient 3D convolution on sparse data.
Multi-view approaches convert the 3D point sets to a collection 
of 2D views so that the popular 2D convolutional operations 
can be applied on the converted data~\cite{multi-view,kalogerakis20173d}. 
As an example, the multi-view CNN~\cite{multi-view} constructs the CNN 
for each view, and a view pooling procedure is used to aggregate the 
extracted features of each view. 
%

{\bfseries Point-based approach}: 
PointNet~\cite{pointnet} is the milestone work for directly 
processing point sets using the deep neural network.
It extracts each point's feature with a shared MLP 
and aggregates them with a symmetric function, 
such as max pooling, which is independent of input order. 
However, PointNet~\cite{pointnet} cannot combine 
the information of neighbor points. 
To address this issue, PointNet++~\cite{pointnet++} 
uses FPS and neighborhood 
query algorithms to sample centroids and their 
neighbor points and then extracts their features 
using a shared MLP and max pooling. 
The feature extraction operations mentioned 
above still do not take the local spatial distribution 
into account as shown in \figref{fig:compare} (a). 
That is, in existing methods, the operations on points at 
different spatial locations use the same weighting factors.
On the contrary, by combining the SDWs with subsequent operations, we can make such process spatially 
variable.

There are some other concurrent point-based approaches to process point 
sets using deep learning, such as
~\cite{sonet,rsnet,kcnet,pointcnn,splatnet}. 
Especially,
SO-Net~\cite{sonet} applies the self-organizing network on the point sets processing.
RSNet~\cite{rsnet} uses Recurrent Neural Network (RNN) 
to process point sets. KCNet~\cite{kcnet} introduces 
the kernel correlation to combine the information of 
the neighborhood. And PointCNN~\cite{pointcnn} 
learns a $\chi$ transform from the point sets 
to permute them in canonical order.
In~\cite{splatnet}, they project 
the point features into regular domains, so that the 
typical CNNs can be applied.
The sparse data can also be represented as meshes
~\cite{monti2017geometric} 
or graphs
~\cite{fast-localized-spectral-filtering,syncspeccnn}, and there are some works that aim at learning feature 
from these representations. We refer the reader 
to~\cite{geometric} for a more comprehensive survey.
%



\vspace{-5pt}
\section{Our Method}
\label{sec:section3}
Firstly, we introduce the method of extracting spatial 
distribution feature of the local region; then the generation 
of Spatial Distribution Weights (SDWs), which based on the spatial distribution 
feature, is described in depth. We elaborate on the 
integration of Spatial Distribution Weights (SDWs) with other operations and introduce our
LSANet finally.
\subsection{Extract spatial distribution feature}
Let the relative coordinate of each point in a local 
region is $\{P_{i}|P_{i}\in \mathbb{R} ^{3},i=1,...,K\}$, 
where $K$ is the number of points in a local region. 
The spatial distribution feature consists of two parts, 
one is the spatial feature of the point itself, 
and the other is the spatial feature of the local region 
where the point is located. 

The spatial feature of the point can be expressed as:
\begin{equation}
    S_{i}^{p} = \boldsymbol{W_{0}}P_{i},
\end{equation}
where $\boldsymbol{W_{0}} \in \mathbb{R} ^{64\times 3}$, 
and $S_{i}^{p} \in \mathbb{R}^{64}$, 
which is the spatial feature of the point itself. 

We use the following formula to encode 
the spatial distribution of the whole local region:

\begin{equation}
    S^{g} = \frac{1}{K}\sum\limits_{i=1}^{K}\boldsymbol{W_{1}}P_{i},
\end{equation}

where $\boldsymbol{W_{1}}\in \mathbb{R} ^{64 \times 3}$.
As shown above,
$S^{g}$ encode spatial information of all points in the local region. 
To preserve permutation invariance, we apply the same weight $\boldsymbol{W_{1}}$ to 
all points in the local region. 

We 
concatenate
the spatial feature of each point 
with the spatial distribution of the region and 
get the final spatial distribution feature:

\begin{equation}
    S_{i} = [S_{i}^{p}, S^{g}],
\end{equation}
where $[,]$ denotes the concatenation operation, $S_{i} \in \mathbb{R}^{128}$
is the spatial distribution feature 
of each point, which is generated by the above formula and associated with not only spatial 
location itself but also all points of the local region, 
encoding the spatial information explicitly. 
Different points in the same local region share the same $S^g$.
We will utilize each point's spatial distribution feature 
to generate SDWs next.

\subsection{Generation of Spatial Distribution Weights (SDWs)}
Suppose the feature of a local region in the $l$-th sub-layer
is $\{X_{i}^{l}|X_{i}^{l}\in \mathbb{R}^{F_{l}},i=1,...,K\}$, where $F_{l}$ denotes the channel of $X_{i}^{l}$ in the $l$-th layer,
$K$ is the number of points in the local region, and $l$ 
is the index of sub-layers in the \emph{LSA} Layer. 

We use the SDWs generator to generate SDWs
for their subsequent feature extraction operations.
The SDWs generator takes spatial distribution 
feature of local region as input, expressed 
in $\{S_{i}|i=1,...,K\}$, where $S_{i} \in R^{128}$, 
$i$ is the index of the neighboring points. Note that $S_{i}$
is related to the spatial structure of $X_{i}^{l}$ and its local region.
In order to generate the first SDWs
for the corresponding operation, we define the 
SDWs as $e_{i}^{1}=f_{\theta}(S_{i})$, 
where $f_{\theta}$ is a non-linear function 
which is determined by the learnable parameters 
$\theta$. 
In this work, we use a fully connected network 
as $f_{\theta}$ to get the first SDWs $e_{i}^{1}$, 
which can be expressed as:
\begin{equation} 
e_{i}^{1} =\sigma(\boldsymbol{W_{s}^1}(S_{i})),
\label{equ:equ1}
\end{equation}
where $\sigma(\cdot)$ denotes the sigmoid function, 
$\boldsymbol{W_{s}}^1\in R^{F_{1}\times 128}$, 
and $s$ means that the $W_{s}$ belongs to our SDWs generator. 
For the output $e_{i}^{1} \in R^{F_{1}}$, 
it has the same dimension as the point feature $X_{i}^{1}$. 
We can use the following formulation to generate new 
SDWs for the further feature learning process:
\begin{equation} 
e_{i}^{l} = \sigma(\boldsymbol{W_{s}^{l-1}}(e_{i}^{l-1})),
\label{equ:equ2}
\end{equation}
where $\boldsymbol{W_{s}^{l-1}}\in R^{F_{l}\times F_{l-1}}$.
Note that $e_{i}^{l} \in R^{F_l}$ shares the same dimension as $X_{i}^{l}$.
We use the $ReLU$ activation function after each $W$ 
to introduce nonlinearity. Therefore, the formula mentioned above generates 
the expected SDWs which are related to the spatial distribution in each local region. 
Note that the process mentioned above can be 
easily extended with multiple local regions.
\figref{fig:attention_sa_module} (b) shows the whole processes.


\subsection{Combine SDWs with other operations}
\label{section3.1.2}
Next, we show how our SDWs participate in other 
feature extraction operations, which allows the feature 
extraction processes to take the local spatial distribution into account.
For example, combining the SDWs with the 
shared MLP can be expressed as follows:

\begin{equation}
X_{i}^{l} = \boldsymbol{W^{l-1}_{m}}(X_{i}^{l-1}\otimes e_{i}^{l-1})= \boldsymbol{W_{m}^{'l-1}}X_{i}^{l-1},
\label{equ:equ3}
\end{equation}
where $\otimes$ denotes element-wise multiplication, 
$\boldsymbol{W^{l-1}_{m}}\in R^{F_{l}\times F_{l-1}}$, the parameter $m$ means 
that the weight belongs to the shared MLP operation 
which is shown in \figref{fig:attention_sa_module} (a),
%
$e_{i}^{l-1}\in R^{F_{l-1}}$, and $X_{i}^{l-1}\in R^{F_{l-1}}$.
As shown in \equref{equ:equ3}, the value of $W^{l-1}_{m}$ 
is independent with the spatial coordinate of $X_{i}^{l-1}$ and 
shared across different points in the local region.
After
combined with the SDWs $e_{i}^{l-1}$, the value of 
the updated weight $W^{'l-1}_{m}$ is related to the 
spatial distribution. 
For each point $X_{i}^{l-1}$ in the local region,  
$\boldsymbol{W^{'l-1}_{m}}$ can adaptively learn to assign 
different weights according to its spatial distribution, 
with which the local shape pattern can be captured better. 
The entire process is shown in \figref{fig:attention_sa_module} (a).
%

The max pooling operation selects the point with the 
strongest response in each channel regardless of spatial 
relationship in the local region. 
However, by combining the SDWs, 
the pooling operation can be guided to select the optimal point 
based on its spatial distribution, which can be formulated as: 
\begin{equation}
Y = \underset{{i\in 1,...,K}}{max}(X_{i}^{l}\otimes e_{i}^{l}),
\label{equ:equ4}
\end{equation}
where $Y \in R^{F_{l}}$, $X_{i}^{l}\in R^{F_{l}}$, and $e_{i}^{l}\in R^{F_{l}}$.
We will further investigate it by experiments to show 
that the combination of the SDWs with the  
feature extraction operations can lead to better results, which is shown in \tabref{tab:module_validation}. 
%


\subsection{LSANet Architecture}

To combine spatial coordinates with the features in each layer better, 
we propose an additional branch architecture, 
in which \emph{Spatial Feature Extractor (SFE)} 
is mounted to get high-dimension spatial 
representation as shown in \figref{fig:pipline}.

The input of \emph{SFE} are the raw coordinates of local regions or the spatial feature 
from the previous \emph{SFE}. 
To improve the dimensions of the coordinates,
we send the spatial coordinates information of input to shared MLP.  
Then we combine the output of the shared MLPs with the input 
and use it as the spatial information 
that flows into the backbone network for abstract representation.
Finally, we use shared MLP to enhance the representation of spatial 
information further and inflow it into the next \emph{SFE}.  
In this way, we can lift the dimension of raw coordinates and get more 
abstract representation layer by layer.

The architecture of LSANet is shown 
in \figref{fig:pipline}. 
Note that we use LSA layer as our basic unit, 
and add the additional branch to enhance the spatial 
feature representation using the \emph{SFE}. 
We use farthest point sampling (FPS) and ball query algorithms to sample and group, which are the 
same as PointNet++.
The output features of the last \emph{LSA} layer are aggregated 
by a fully connected network for classification. 
The segmentation model extends the classification model using the FP module 
in PointNet++~\cite{pointnet++} to upsample the reduced points and outputs 
per-point scores for semantic labels.

\section{Experiments}
\label{sec:section4}
 
We evaluate the performance of the proposed \emph{LSA} Layer and 
LSANet with extensive experiments.
First, the experimental results of our LSANet  
and other \sArt point-based approaches on the ModelNet40~\cite{wu20153d}, ShapeNet~\cite{yi2016scalable}, ScanNet~\cite{dai2017scannet}, and S3DIS~\cite{armeni20163d} are shown in \secref{section4.1}.
Second, we perform the ablation study to 
validate our LSANet design, and then visualize what our 
\emph{LSA} layer learns in \secref{section4.2}.
At last, we analyze the space and time complexity 
in \secref{section4.3}. 


\begin{table*}[tp]  
\centering
  \small
  \renewcommand{\tabcolsep}{2.25mm}

    \begin{tabular}{c|c c|c c| c c|c|c c}  
    \hline
    Task&\multicolumn{4}{c|}{Classification}&\multicolumn{5}{c}{Segmentation}\\
    \hline
    \multirow{2}{*}{Dataset}&\multicolumn{4}{c|}{ModelNet40}&\multicolumn{2}{c|}{\multirow{2}{*}{ShapeNet}} & \multirow{2}{*}{ScanNet} & \multicolumn{2}{c}{\multirow{2}{*}{S3DIS}} \\
    \cline{2-5}
           &\multicolumn{2}{c|}{Pre-aligned}&\multicolumn{2}{c|}{Unaligned} &      &      &      & \\
    \hline
    Metric &mA    &OA     &mA    &OA     &mpIOU  &pIOU  &OA    &OA    &mIoU  \\
    \hline
    KCNet~\cite{kcnet}                &-&91.0\%&-      &-  &82.2\% &84.7\%  &-     &-     &-     \\
    Kd-Net\cite{kdtree}               &88.5\% &91.8\%  &-&-&77.4\% &82.3\%  &-     &-     &-     \\
    SO-Net~\cite{sonet}               &-      &90.9\%  &-&-&81.0\% &84.9\%  &-     &-     &-     \\
    PCNN~\cite{pcnn}                  &-      &92.3\%  &-&-&81.8\% &85.1\%  &-     &-     &-     \\
    SPLATNet\cite{splatnet}           &-      &-       &-&-&83.7\% &85.4\%  &-     &-     &-     \\
    SpecGCN~\cite{spectral_graph_conv}&-&-&-      &91.5\%  &-      &85.4\%  &84.8\%&-     &-     \\
    SpiderCNN~\cite{spidercnn}        &-&-&-      &90.5\%  &81.7\% &85.3\%  &81.7\%&-     &-     \\
    SCN~\cite{ShapeContextNet}        &87.6\% &90.0\%  &-      &84.6\%  &-     &-     &-     \\
    PointNet~\cite{pointnet}          &-&-&86.2\% &89.2\%  &80.4\% &83.7\%  &-     &78.5\%&47.6\%\\
    PointNet++~\cite{pointnet++}      &-&-&-      &90.7\%  &81.9\% &85.1\%  &84.5\%&-     &-     \\
    SyncSpecCNN~\cite{syncspeccnn}    &-&-&-      &-       &82.0\% &84.8\% &-     &-     &-         \\
    PointCNN~\cite{pointcnn}          &88.8\%&92.5\%&88.1\%  &92.2\% &84.6\% &86.1\%   &85.1\%     &85.14\%  &65.39\% \\
    RSNet~\cite{rsnet}                &-&-&-      &-       &81.4\% &84.9\%  &-     &-     &56.5\%\\
    SPG~\cite{spg}                    &-&-&-      &-       &-      &-       &-     &85.5\%  &62.1\% \\

    \hline
    LSANet (ours)                              &\textbf{\color{black}{90.3\%}} &\textbf{\color{black}{93.2\%}} &\textbf{\color{black}{89.2\%}}&\textbf{\color{black}{92.3\%}}&\textbf{\color{black}{83.2\%}} &\textbf{\color{black}{85.6\%}} &\textbf{\color{black}{85.1\%}}&\textbf{\color{black}{86.8\%}} &\textbf{\color{black}{62.2\%}} \\

    \hline  
    \end{tabular}  
  \caption{Comparisons with other point-based networks on ModelNet40~\protect\cite{wu20153d} in per-class accuracy (mA) and overall accuracy (OA), ShapeNet~\protect\cite{yi2016scalable} in part-averaged IoU (pIoU) and mean per-class pIoU (mpIoU), Scannet~\protect\cite{dai2017scannet} in per voxel overall accuracy (OA), and S3DIS~\protect\cite{armeni20163d} in mean per-class IoU (mIoU) and overall accuracy(OA).}  
  \label{tab:performance_comparison}  
  \vspace{-10pt}
\end{table*}

\subsection{Classification and Segmentation Tasks}
\label{section4.1}
{\bfseries Dataset}: We apply our LSANet on the following datasets:
\begin{itemize}
    \item 
    ModelNet40~\cite{wu20153d}: This dataset includes 12,311 CAD models 
    from the 40 categories, and we use the official split with 9,843 for training 
    and 2,468 for testing. To get the 3D points, we sample 1,024 points uniformly 
    from the mesh model.
    \item
    ShapeNet~\cite{yi2016scalable}: 6,880 models from 16 shape categories 
    and 50 different parts consist in the ShapeNet~\cite{yi2016scalable}, and 
    each shape is annotated with 2 to 6 parts. Following~\cite{pointnet++}, 
    we use 14,006 models for training and 2,874 for testing, 2,048 points are 
    sampled uniformly from each CAD models, and each point is associated with 
    a part label. These points with their surface normals are used as input, 
    assuming that the category labels are known.
    \item
    ScanNet~\cite{dai2017scannet}: The ScanNet~\cite{dai2017scannet} 
    is a large-scale semantic segmentation dataset containing 2.5M views in 1513 scenes.
    Since ScanNet is constructed from real-world 3D scans of indoor scenes,
    it is more challenging than the synthesized 3D datasets. In our experiment,
    we follow the configuration in~\cite{pointnet++} and use 1201 scenes for training, 
    312 scenes for testing with 8192 points as our inputs. We remove the RGB information 
    in this experiments and only use the spatial coordinates as input. 
    \item
    S3DIS~\cite{armeni20163d}: The S3DIS dataset contains 3D scans in 6 areas including 271 rooms. 
    Each point is annotated with the label from 13 categories. We follow the way in ~\cite{pointnet} to prepare training data and split the training and testing set with k-fold strategy. 8192 points are sampled in each block randomly for training. We use XYZ, RGB and normalized location on each point as input.
\end{itemize}
%


{\bfseries Network Configuration}: 
The configuration of LSANet is shown in \tabref{tab:architecture}.
We use Adam optimizer, and the initial 
learning rate is 0.002 which is applied with exponential decay. 
The decay ratio is 0.7 applied every 40 epochs. 
We use the ReLU activation function. Batch normalization is applied after each MLP, 
and the batch size of data is 32. 
We train the LSANet for 250 epochs on two NVIDIA GTX 1080Ti GPUs.

%
%

\begin{table}[tb]
    \centering
  \small
  \renewcommand{\arraystretch}{1.6}
  \renewcommand{\tabcolsep}{0.2mm}
  \vspace{-7pt}
    \begin{tabular}{c c |c c c c}
    \hline
         Datasets          &  & L1 & L2 & L3 & L4 \\
    \hline
    \hline
\multirow{3}{*}{\begin{sideways}ModelNet40\end{sideways}}& $N$&  512  & 128   & 1   &  -  \\
                           & $K$&  32  &  64  & 128   &  -  \\
                           & $F$& (64,64,128)   & (128,128,256)  &(256,512,1024)&\\
    \hline

\multirow{3}{*}{\begin{sideways}ShapeNet\end{sideways}}& $N$&  512  & 128   & 1   &  -  \\
                            & $K$&  64  &  64  & 128   &  -  \\
                           & $F$& (64,64,128)   & (128,128,256)  &(256,512,1024)&\\
    \hline

\multirow{3}{*}{\begin{sideways}ScanNet\end{sideways}}& $N$&  1024  & 256   & 64   &  16  \\
                           & $K$&  32  &  32  & 32   &  32  \\
                           & $F$& (32,32,64)   & (64,64,128)  &(128,128,256)& (256,256,512)\\
    \hline

\multirow{3}{*}{\begin{sideways}S3DIS\end{sideways}}& $N$&  1024  & 256   & 64   &  16  \\
                        & $K$&  32  &  32  & 32   &  32  \\
                           & $F$& (32,32,64)   & (64,64,128)  &(128,128,256)& (256,256,512)\\
    \hline
    \end{tabular}
    \caption{The backbone architecture of our LSANet for each dataset. In each \emph{LSA} layer, $N$ stands for the number of local regions, $K$ represents the number of points in each local region, and $F$ stands for the output dimensions of shared MLP in \emph{LSA} layer.}
    \label{tab:architecture}
    \vspace{-15pt}
\end{table}

\begin{figure*}[t]
    \centering
    \begin{overpic}[width=.95\textwidth]{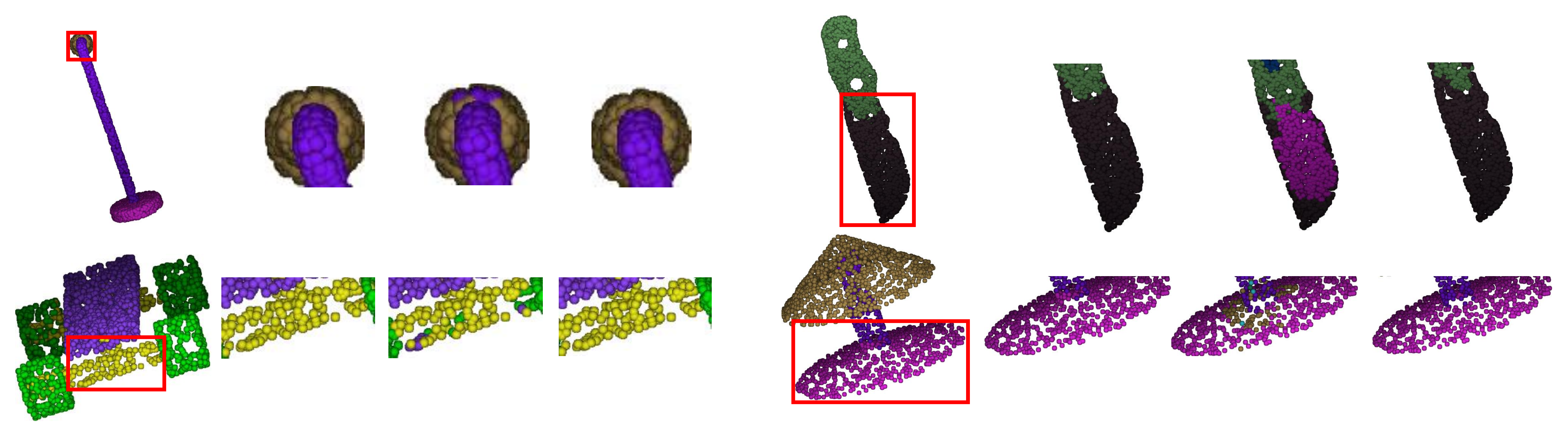}
    \put(17,0){GT}
    \put(25,0){PointNet++}
    \put(38,0){Ours}
    
    \put(68,0){GT}
    \put(76,0){PointNet++}
    \put(93,0){Ours}
    \end{overpic}
    \vspace{-5pt}
    \caption{The visualizations on ShapeNet.} 
    \label{fig:ShapeNet_vis}
\end{figure*}

\begin{figure}[t]
    \centering
    \vspace{-20pt}
          \includegraphics[width=8.5cm]{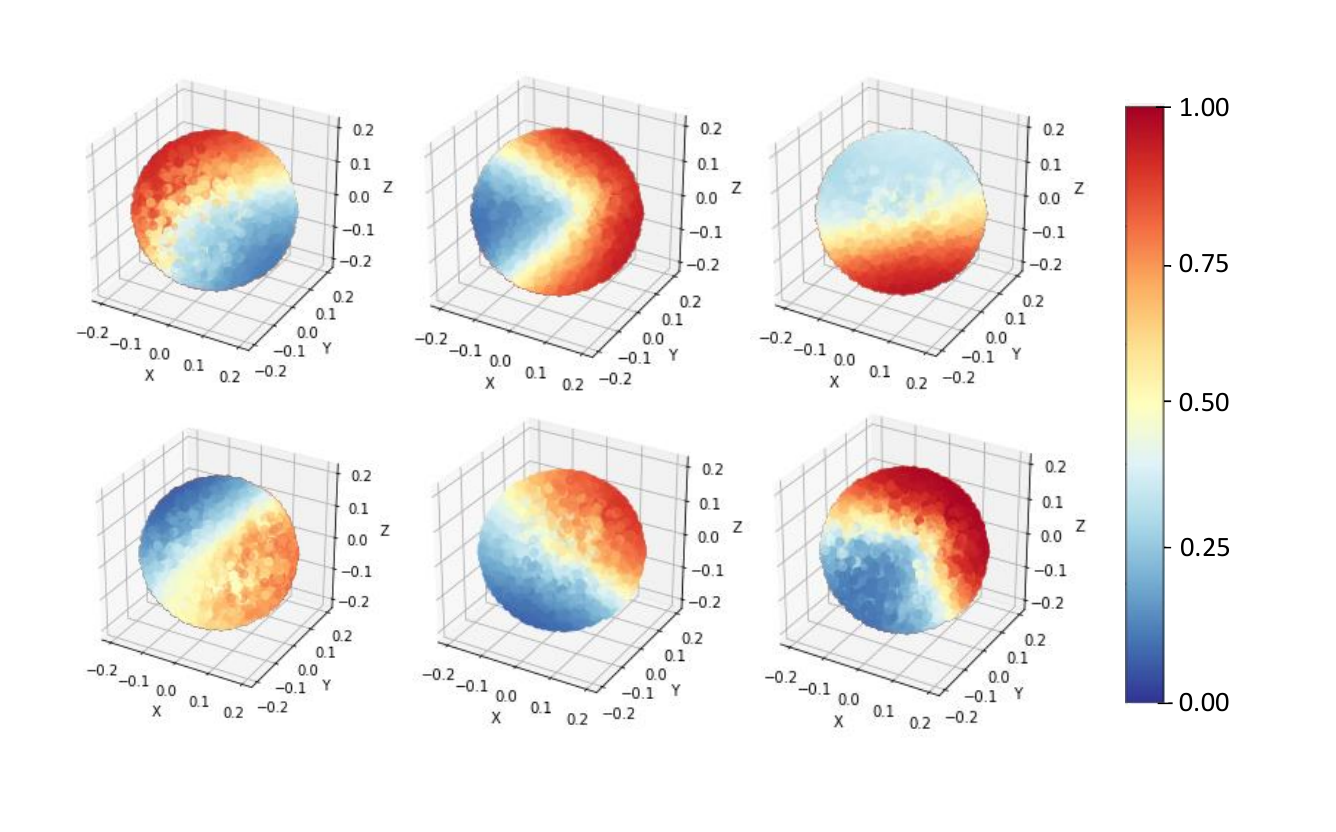}
    \vspace{-35pt}
    \caption{
    {\bfseries Visualization of the SDWs}:
We visualize the channel-wise SDWs of \emph{LSA} 
layer to these local regions' spatial coordinates before 
shared MLP operation. In this figure, we randomly 
sample 6 channels from 64 feature channels to show. 
It can be observed that our SDWs are 
spatially related in each feature channel.} 
    \label{fig:kernel_attention_vis}
\end{figure}

{\bfseries Results}: \tabref{tab:performance_comparison} compares our  
results with \sArt works on the datasets mentioned above.
For the task of Classification,
We divide the settings into the pre-aligned and the unaligned according to 
whether they rotate randomly during the training or testing phase, due to a large portion
of the 3D models from ModelNet40 are pre-aligned. To compare fairly, we report our LSANet's performance
in both settings.
We use the overall accuracy as the evaluation metric. 
For the input of 1024 points without surface normal, 
in terms of the overall accuracy and Unaligned setting, 
our method achieves 1.6\% higher than the multi-scale grouping (MSG) network 
of PointNet++ even though we do not use multi-scale grouping (MSG) in the LSA layer.
Our LSANet also outperforms the 
PointNet++'s MSG architecture 
which uses both 5000 points 
and surface normal as input.
These results show the effectiveness of our LSA Layer,
and in general, we realize better accuracy than other 
methods in both settings.  In the segmentation task, we evaluate our LSANet on the ShapeNet, ScanNet,and S3DIS.  We note that
our method outperforms all the compared methods, such as PointNet++ which does not have our 
\emph{LSA} layer and additional branch.
Our LSANet also outperforms the approaches based on~\cite{pointnet} such as SpecGCN~\cite{spectral_graph_conv} and SpiderCNN~\cite{spidercnn}. The visualizations of segmentation results on ShapeNets 
are
shown 
in \figref{fig:ShapeNet_vis}.


\subsection{Analysis and Visualization}
\label{section4.2}
We now validate our proposed LSANet design by control experiments with classification task on the ModelNet40~\cite{wu20153d} dataset under unaligned settings,
and then we visualize the SDWs generated by our \emph{LSA} module.
%

{\bfseries Module validation:} We demonstrate the positive effects 
of our \emph{LSA} Layer and network architecture by ablation experiment.  We also remove the 
integration of SDWs from max pooling and the region spatial encoder
part of \emph{LSA} Layer in \figref{fig:attention_sa_module} to verify their effectiveness.
The detailed results are shown in \tabref{tab:module_validation}. 

As shown in these experiments, the \emph{LSA} Layer and \emph{SFE}
bring 1.1\% and 
0.8\%
accuracy improvement respectively,
illustrating the effectiveness of our design. We also observe that the 
region spatial encoder of \emph{LSA} module improves the results, which shows 
the validity of the whole region information. The results also show that 
the max pooling combined with SDWs can select the optimal point based on 
its spatial distribution and achieve better effects.

%


\begin{table}[t]
\small
\centering
\begin{tabular}{l c} 
\hline
 Network Configuration & OA   \\ 
 \hline
baseline  &  90.6\% \\
baseline + \emph{SFE}   &91.4\%  \\
baseline + \emph{LSA}(w/o region spatial encoder) &91.5\% \\
baseline + \emph{LSA}(w/o max-pooling) &91.4\%  \\
baseline + \emph{LSA}                  &91.7\% \\
\hline
baseline + \emph{LSA} + \emph{SFE} (ours)     &\textbf{\color{black}{92.3\%}} \\
 \hline
\end{tabular}
\vspace{-5pt}
\caption{ 
Ablation study on ModelNet40 classification task under unaligned settings. 
}

\label{tab:module_validation} 
\vspace{-15pt}
\end{table}

\begin{figure}[t]
 \vspace{-15pt}
    \centering
    \begin{overpic}[width=9cm]{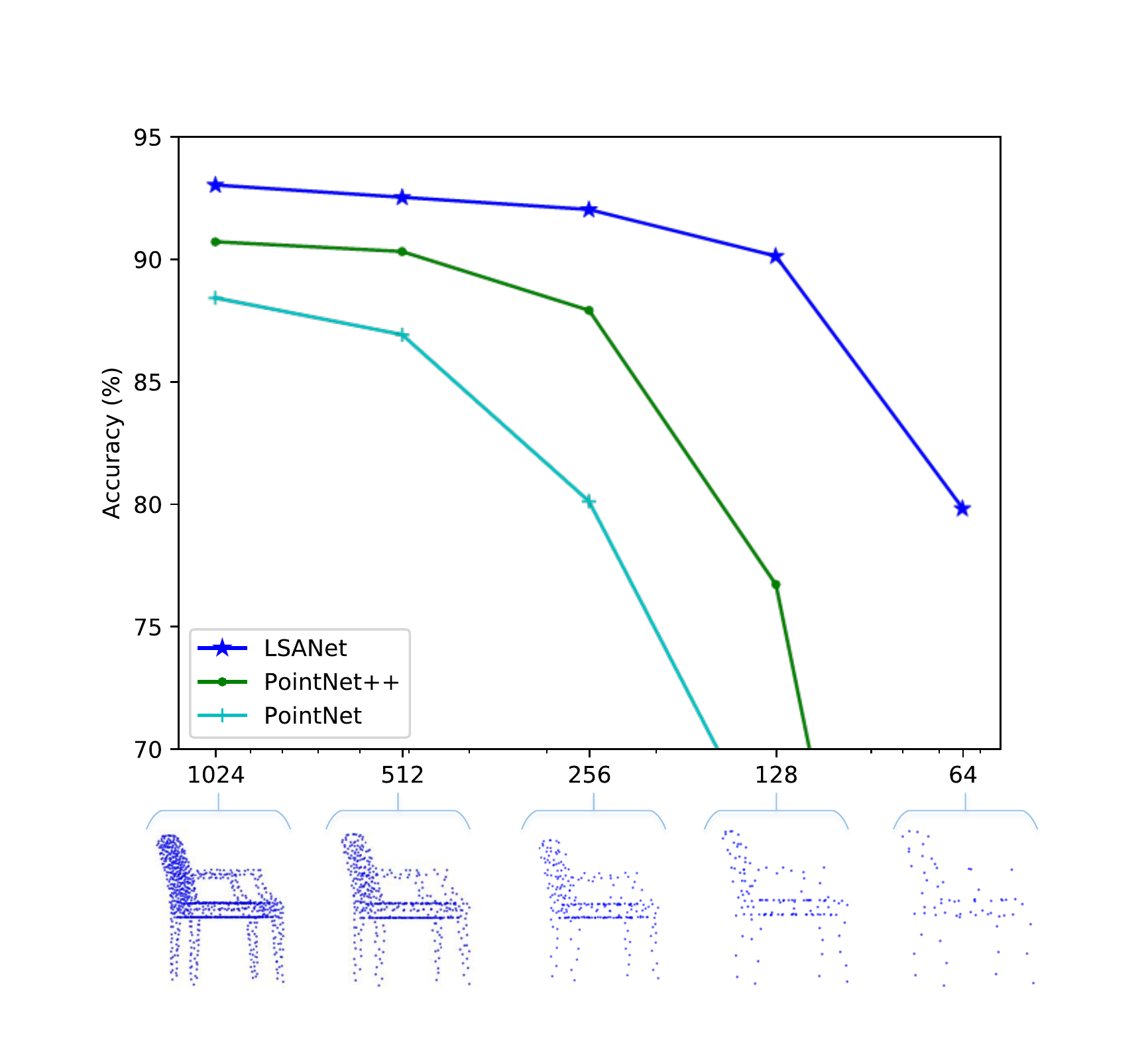}
    \end{overpic}
    \vspace{-20pt}
    \caption{
    The test results of using
    different number of points as input to the same model trained with 1024 points.
    }
    \label{fig:density}
    
    \vspace{-15pt}
\end{figure}

{\bfseries Sampling density :}
We test the robustness of our LSANet to sampling density by 
using 1024, 512, 256,128 and 64 points sampled on ModelNet40 dataset as 
input. Our LSANet is trained on ModelNet40 dataset using 1024 points. 
We use random input dropout in for a fair comparison. The test results are shown in 
\figref{fig:density}. We compare our LSANet with PointNet~\cite{pointnet} and PointNet++~\cite{pointnet++}.
We can see that as the number of points decreases, it becomes more and more difficult for the people to judge.
But our LSANet performs well at different number of points.

{\bfseries Visualization of the SDWs:}
In \figref{fig:kernel_attention_vis}, 
we randomly pick 512 representative points with 
their neighboring ones of an object in the 
test set of ModelNet40~\cite{wu20153d} dataset, 
and visualize 
the response of \emph{LSA} Layer to these local 
regions before MLP in each 
channel (as discussed in \secref{section3.1.2}). 
It is obvious to see that our SDWs 
obtain
different 
preferences for directions in each channel. 
This module guarantees that our LSANet 
can effectively perceive fine-grained patterns 
by learning SDWs.
%
\subsection{Complexity Analysis}
 \label{section4.3}
We further compare both space and time complexities with other methods, 
in which the classification network is used.
\tabref{tab:model_param1} shows that our LSANet has proper parameters with 
fast inference time.
In addition, our segmentation network involves fewer parameters 
than our classification network (see \tabref{tab:model_param2}).

\begin{table}[ht]
\small
\centering
\vspace{-10pt}
  \renewcommand{\tabcolsep}{0.6mm}
\begin{tabular}{l c c} 
\hline
 Method  &  Parameters & Inference time \\
\hline
 PointNet++ (SSG)~\cite{pointnet++} &  1.48M  & 0.027s  \\
 SpecGCN~\cite{spectral_graph_conv}&   2.05M  & 11.254s\\
 SpiderCNN~\cite{spidercnn}        &   5.84M  & 0.085s \\
 \hline
 LSANet (ours)                    &   2.30M  & 0.060s  \\
 \hline

\end{tabular}
\vspace{-5pt}
\caption{
Comparison of different methods on the number of parameters and inference time for Classification task.}
\label{tab:model_param1} 
\vspace{-10pt}
\end{table}

\begin{table}[h]
\small
\centering
\vspace{-5pt}
\begin{tabular}{l c c} 

\hline
 Datasets   &               Task     & Parameters  \\
\hline
 Modelnet40~\cite{wu20153d} &       Classification    &      2.30M\\
 ShapeNet~\cite{yi2016scalable}   &       Segmentation  &      2.24M\\
 ScanNet~\cite{dai2017scannet}    &   Segmentation  &      1.36M\\
 S3DIS~\cite{armeni20163d}      &   Segmentation  &      1.41M\\
 \hline
\end{tabular}

\caption{The number of our LSANet's parameters  on 
four datasets.}
\label{tab:model_param2} 

\end{table}

\section{Conclusion}

In this work, we propose a novel \emph{LSA} Layer and LSANet.
Based on such new design, our LSANet has more powerful 
spatial information extraction capabilities 
and provides on par or better results 
than \sArt approaches on standard benchmarks for different
3D recognition tasks including object classification, 
part segmentation, and semantic segmentation. 
We also provide ablation experiments and 
visualizations to illustrate the effectiveness 
of our LSANet design.
\bibliographystyle{named}
\bibliography{ijcai19}

\end{document}